\documentclass[runningheads]{llncs}
\usepackage{graphicx}
\usepackage{multirow}
\usepackage[T1]{fontenc}
\usepackage{mathbbol}

\begin{document}

\title{Damage GAN: A Generative Model for Imbalanced Data}

\author{Ali Anaissi\inst{1}\and
Yuanzhe	Jia\inst{1}\and
Ali	Braytee\inst{2}\and
Mohamad	Naji\inst{2} \and
Widad Alyassine \inst{1}
}

\authorrunning{Anaissi et al.}

\institute{School of Computer Science, The University of Sydney \and
School of Computer Science, The University of Technology Sydney
\email{ali.anaissi@sydney.edu.au, yjia5612@uni.sydney.edu.au, ali.braytee@uts.edu.au, Mohamad.Naji@uts.edu.au, widad.yassien@gmail.com}}

\maketitle

\begin{abstract}
This study delves into the application of Generative Adversarial Networks (GANs) within the context of imbalanced datasets. Our primary aim is to enhance the performance and stability of GANs in such datasets. In pursuit of this objective, we introduce a novel network architecture known as Damage GAN, building upon the ContraD GAN framework which seamlessly integrates GANs and contrastive learning. Through the utilization of contrastive learning, the discriminator is trained to develop an unsupervised representation capable of distinguishing all provided samples. Our approach draws inspiration from the straightforward framework for contrastive learning of visual representations (SimCLR), leading to the formulation of a distinctive loss function. We also explore the implementation of self-damaging contrastive learning (SDCLR) to further enhance the optimization of the ContraD GAN model. Comparative evaluations against baseline models including the deep convolutional GAN (DCGAN) and ContraD GAN demonstrate the evident superiority of our proposed model, Damage GAN, in terms of generated image distribution, model stability, and image quality when applied to imbalanced datasets.

\end{abstract}

\section{Introduction}

GAN is a popular deep learning architecture composed of a generator and a discriminator. The generator aims to learn the distribution of real samples, while the discriminator evaluates the authenticity of inputs, creating a dynamic "game process". While GAN is extensively used for image generation, their effectiveness in imbalanced datasets has received less attention\cite{borji2022pros,naji2021anomaly,anaissi2023b}. Contrastive learning, a self-supervised training technique that captures augmented image invariants and reduces training effort for image classification, has recently gained prominence \cite{anaissi2023multi}. Our objective is to integrate contrastive learning into GAN, leveraging its potential to improve the performance and stability in the context of imbalanced datasets.

In this paper, we introduce Damage GAN, a novel GAN model constructed by implementing SDCLR as a replacement for the SimCLR module in ContraD GAN, and applying it to the task of training and validation. The imbalanced CIFAR-10 dataset is utilized for both model training and validation. Evaluation metrics such as Fréchet Inception Distance (FID) and Inception Score (IS) are employed to assess the performance of the proposed model. After experiments, we demonstrate that Damage GAN outperforms state-of-the-art models, such as DCGAN and ContraD GAN, when applied to imbalanced datasets, thereby highlighting its potential for improving GAN performance on imbalanced datasets.

This paper is organized as follows. Section \ref{sec:relatedwork} is about related work, mainly discusses and analyses the contributions and limitations related to our study in the last ten years. 
In section \ref{sec:method} we describe our research ideas and model structure in details. In section \ref{sec:experiment} we clarify the dataset used for experiments, the model training procedure, the evaluation metrics, as well as how to conduct experiments on exploratory data analysis and model comparison to verify the predictive ability, stability and applicability of our model. Finally, we conclude the paper in section \ref{sec:conclusion} and look to the future. 
Since this paper contains a large number of technical terms, the notations are summarized in Table \ref{tab:notations} for the convenience.

\begin{table}
\caption{Notations used in this paper.}
\label{tab:notations}
\centering
\begin{tabular}{ p{3cm} p{8cm} }
\hline
Abbreviation & Description \\
\hline

CIFAR-10 & It is an image database containing 60,000 32x32 colour images in 10 classes, with 6,000 images per class.\\
ContraD GAN & Training GANs with stronger augmentations via contrastive discriminator.\\
CNN & Convolutional Neural Network.\\
CV & Computer Vision.\\
Damage GAN & Proposed model in this paper.\\
DCGAN & Deep Convolutional Generative Adversarial Network.\\
FID & Fréchet Inception Distance.\\
ImageNet & It is an image database organized according to the WordNet hierarchy, in which each node of the hierarchy is depicted by hundreds and thousands of images.\\
IS & Inception Score.\\
GAN & Generative Adversarial Network.\\
LeakyReLU & It is a type of activation function based on ReLU. It has a small slope for negative values with which LeakyReLU can produce small, non-zero, and constant gradients with respect to the negative values.\\
MLP & Multi-Layer Perceptron.\\
NLP & Nature Language Processing.\\
ReLU & Rectified Linear Units. It is a non-linear activation function that is widely used in multi-layer neural networks or deep neural networks.\\
SDCLR & Self-damaging Contrastive Learning of Visual Representations.\\
Sigmoid & It is a special form of the logistic function with an S-shaped curve.\\ 
SimCLR & Simple Framework for Contrastive Learning of Visual Representations.\\
SoftMax & It is a function that turns a vector of $K$ real values into a vector of $K$ real values that sum to 1.\\
Tanh & Hyperbolic Tangent. It is the hyperbolic analogue of the Tan circular function used throughout trigonometry.\\

\hline
\end{tabular}
\end{table}

\section{Related Work}
\label{sec:relatedwork}

The foundation of our proposed model is built upon the ContraD GAN, which involves training a generative adversarial network through contrastive learning applied to the discriminator. To further improve the model's performance, we recommend incorporating Self-Damaging Contrastive Learning (SDCLR) as a replacement for the SimCLR module within the ContraD GAN framework.

The following section introduces a literature review covering fundamental theories pertinent to our proposed model.

\subsection{GAN}

The framework of GAN, which is a type of generative algorithm, was proposed in 2014\cite{goodfellow2014explaining}. The main idea of the generative model is to learn the pattern of the training data and use that knowledge to create new examples\cite{gui2021review,zhou2022vgg,yao2022conditional}. GAN introduces the concept of adversarial learning to address the limitations of generative algorithms. The basic principle is to produce data that looks very similar to real samples. GAN consists of two modules: a generator and a discriminator, which are typically implemented using neural networks. The generator learns to understand the distribution of real examples and generate new ones, while the discriminator tries to determine if the inputs are genuine or fake. The goal is for the generator to capture the true distribution of real data and generate realistic examples.

In recent years, GANs and their variations have gained widespread use in the fields of Computer Vision (CV) and Natural Language Processing (NLP). These models provide distinct advantages over other generative methods, as discussed by Goodfellow et al. in their seminal work\cite{goodfellow2016nips}. GANs offer parallel generation capabilities, a unique feature not present in other generative models. Unlike models like Boltzmann machines, GANs impose fewer restrictions on the generator structure and eliminate the need for Markov chains. Additionally, GANs are recognized for their ability to produce high-quality samples that often outperform those generated by alternative algorithms.

However, GANs also exhibit certain limitations that necessitate consideration. GAN training involves achieving Nash Equilibrium, a complex task that presents challenges. The training process for GANs can be hindered by issues such as oscillation and non-convergence. Partial mode collapses within the generator can result in a limited variety of generated outputs. Imbalances between the generator and discriminator can lead to overfitting concerns.

\subsection{CNN}

\begin{figure}[!t]
\centerline{\includegraphics[scale=0.4]{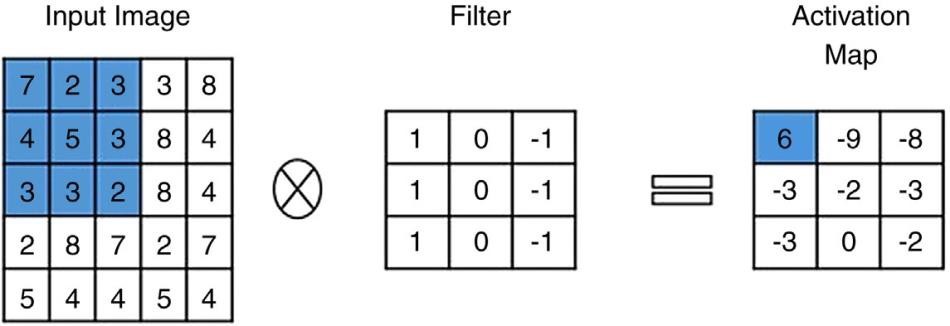}}
\caption{An example of convolutional layer.}
\label{fig:cnn}
\end{figure}

CNNs are a type of neural network that utilizes convolutional filters to capture features in a grid-like manner, resembling the structure of the visual cortex in the human brain. This approach requires less data pre-processing compared to traditional neural networks, as demonstrated by Heaton\cite{heaton2018ian}. Notably, CNNs have become dominant in the field of Computer Vision (CV) with notable works such as LeNet-5\cite{lecun1998gradient}, AlexNet\cite{krizhevsky2012imagenet}, VGG\cite{simonyan2014very}, Inception\cite{szegedy2014scalable}, and ResNet\cite{he2015delving}.
Convolutional Neural Networks (CNNs) present a range of advantages compared to traditional neural networks. Firstly, CNNs excel in reducing the number of specified parameters, contributing to enhanced generalization and a reduced risk of overfitting. Additionally, the architecture of CNNs enables them to learn intricate features from input data via convolutional and pooling layers. Simultaneously, they efficiently carry out classification tasks using fully connected layers, ensuring an organized approach to information processing. Furthermore, CNNs simplify the process of implementing large-scale networks, making them a preferred choice for handling complex tasks in various domains.

The CNN architecture comprises different layers, which are outlined below.
\textbf{Convolutional Layer} effectively captures the features of the input image by using convolutional filters, also known as kernels, to create N-dimensional activation maps. Several hyper-parameters need to be determined and optimized in this process, including the number of kernels, kernel size, activation function, stride, and padding. 
In Figure \ref{fig:cnn}, for example, a 3x3 filter is applied to the input image, moving with a step equal to the chosen stride. At each position, the filter's metrics are multiplied with the corresponding 3x3 spatial elements of the input image. 
By performing this dot product operation for the activation map can be generated.
\textbf{Pooling Layer} follows the convolutional layer in order to decrease the size of feature maps and lower the dimensionality of the network. Prior to this process, the stride and kernel size need to be manually determined. Several pooling methods can be employed, such as average pooling, min pooling, max pooling and mixed pooling.
\textbf{Fully Connected Layer} is positioned as the last layer in the CNN structure and functions as a classifier. In this layer, each neuron is connected to every neuron in the preceding layer. Once the convolutional layers extract features and the pooling layers down-sample the outputs, the resulting outputs are passed through the fully connected layer to generate the final outputs of CNN. It is essential to activate the fully connected layer using a non-linear function like SoftMax, Tanh, or ReLU\cite{sharma2017activation}.

\subsection{DCGAN}

\begin{figure}[b]
\centerline{\includegraphics[scale=0.6]{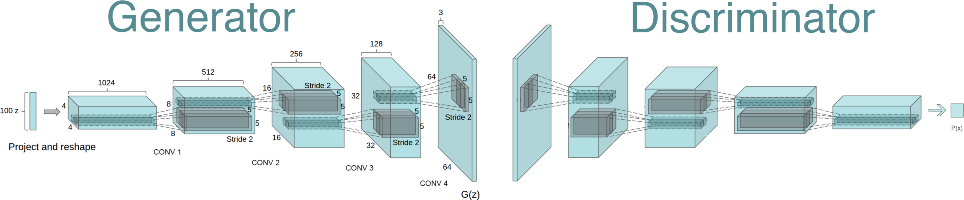}}
\caption{The structure of DCGAN.}
\label{fig:dcgan}
\end{figure}

Unsupervised learning through CNNs garnered noteworthy attention in 2015. A notable milestone was the inception of DCGAN, which showcased the prowess of CNNs in producing visually captivating outcomes\cite{radford2015unsupervised}.
As depicted in Figure \ref{fig:dcgan}, the structure of DCGAN aligns with the foundational architecture of the original GAN. Nonetheless, DCGAN introduces certain alterations. Notably, the generator in DCGAN generates 100-dimensional noise, subsequently subjecting it to processing and transformation via convolutional layers.

For the effective integration of deep convolutional networks within GANs, a series of architectural principles have been introduced. Firstly, the discriminator replaces pooling layers with stride convolutions, enabling spatial down-sampling. In contrast, the generator employs fractional stride convolutions for spatial up-sampling. Batch normalization is integrated into every layer of both the generator and discriminator, stabilizing training, mitigating initialization issues, and promoting gradient flow to deeper layers. The adoption of deeper architectures in place of fully connected layers accelerates convergence. The output layer employs the Tanh activation function, while the ReLU activation function is used in other layers, effectively addressing saturation and covering the color space during training. LeakyReLU activation is applied across all discriminator layers, facilitating higher-resolution modeling.

DCGAN has emerged as a more stable GAN training framework, showcasing the ability to learn meaningful image representations in both supervised learning and generative modeling contexts. Despite these advancements, certain challenges persist in practice, including filter collapse and oscillating behavior, which require continued attention..

\subsection{Contrastive Learning}
Traditional supervised learning techniques heavily rely on annotated data, which can pose challenges when dealing with limited annotations. To address the issue of insufficient labeled data, researchers have explored alternative approaches.
Self-supervised learning, a type of unsupervised learning, has gained attention for its ability to create pseudo-labels autonomously through models\cite{liu2021understand}. This empowers the training of unannotated datasets using a supervised framework, effectively mitigating annotation scarcity.

Among the various methods within self-supervised learning, contrastive learning, introduced in 2021\cite{jaiswal2020survey}, stands out. It hinges on a multitude of negative samples and compares distinct samples to produce high-quality outcomes. This technique aims to identify both similarities and differences within a dataset, effectively categorizing data based on these attributes. Contrastive learning's primary objective is to map similar sample representations in close proximity within the embedding space, while ensuring that dissimilar representations are distanced from each other\cite{shim2021active}. This results in the aggregation of positive sample representations and the separation of negative pairs' representations.

In the subsequent discussion, two prominent contrastive learning frameworks, SimCLR and SDCLR, will be elaborated upon.

\subsubsection{SimCLR}

Despite showing respectable performance, various self-supervised learning methods consistently fall short of achieving results comparable to supervised learning\cite{chowdhury2021applying}. However, SimCLR, a straightforward framework that employs self-supervised contrastive learning, manages to outperform supervised learning outcomes, particularly when applied to the ImageNet dataset\cite{chen2020simple}. SimCLR operates based on three core components:

\begin{itemize}
\item Data Augmentation: This process begins by randomly sampling a batch of images, to which two distinct data augmentations are applied.

\item Base Encoder: Utilizing ResNet-50, the encoder extracts vectors from the augmented images and generates representations through a pooling layer.

\item Projection Head: Introducing a non-linear projection, typically in the form of a single-layer MLP. The loss function integrates two primary elements: cosine similarity and loss calculation.
\end{itemize}

The noticeable difference between this framework and conventional supervised learning mainly stems from the chosen data augmentation, the incorporation of a non-linear projection, and the design of the loss function. Our proposed model draws inspiration from the principles of SimCLR, as we strive to further optimize it.

\subsubsection{SDCLR}
To tackle the challenge of data imbalance, specifically the "long tail" problem, a technique known as self-damaging contrastive learning (SDCLR) has emerged to bolster the efficacy of data training\cite{jiang2021self}.
This method employs two networks: the target model, which undergoes training, and the self-competitor, which acts as a pruning mechanism, ensuring consistency between the outputs of both models.

The core principle behind SDCLR is to foster representations capable of capturing nuanced differences among similar data samples. This is achieved by introducing a self-damaging mechanism during the training process. This mechanism penalizes representations that exhibit excessive similarity, even when they pertain to different views of the same data sample.

Through the implementation of SDCLR, we can fortify the robustness of data and effectively confront the complexities linked with data imbalance and the "long tail" phenomenon. Consequently, we can integrate the insights derived from this model into our own approach.

\subsection{ContraD GAN}

ContraD GAN represents a cutting-edge approach that seamlessly combines contrastive learning and Generative Adversarial Networks (GANs)\cite{jeong2021training}. This innovative method introduces a unique strategy by integrating the loss function from SimCLR with contrastive learning techniques. This integration empowers the discriminator to undergo training with heightened data augmentation, resulting in improved model stability and a reduced risk of discriminator overfitting.

The ContraD GAN workflow encompasses several distinct steps. Initially, data augmentation is applied to real data, generating two distinct perspectives. Concurrently, data augmentation is executed on the data produced by the GAN's generator. These viewpoints are then presented to the discriminator, yielding corresponding representations. A projection head is subsequently employed to derive corresponding vectors for these representations.

ContraD GAN has demonstrated its prowess as a successful fusion of GANs and contrastive learning, yielding impressive outcomes across a range of widely used public datasets. A primary advantage lies in its ability to train adversarial networks with enhanced data augmentation. However, it's important to note that the original work mainly emphasizes the amalgamation of SimCLR and supervised contrastive learning, with limited focus on GAN architecture design.

Consequently, the ContraD GAN framework offers substantial room for further enhancements. For instance, addressing data imbalance challenges and extending the applicability of ContraD GAN to various dataset types are promising areas that warrant exploration and expansion.

\section{Methodology - Damage GAN}
\label{sec:method}
The initial version of ContraD GAN introduced an innovative incorporation of SimCLR into the generator, resulting in a distinctive amalgamation of contrastive learning and GAN. This merger brought forth several benefits, including enhanced performance under rigorous data augmentation and the mitigation of challenges like overfitting. However, the efficacy of ContraD GAN, akin to numerous contrastive learning models, is susceptible to the characteristics of the input dataset.

ContraD GAN has showcased commendable outcomes for classification tasks when operating with sizable, balanced datasets. Nonetheless, complexities arise when confronted with diminutive, highly imbalanced datasets, particularly when classifying items from rare categories characterized by limited representation. Previous research\cite{jiang2021self} has indicated that while contrastive learning exhibits greater resilience to imbalanced data in comparison to supervised learning, it still encounters hurdles when addressing imbalances within long-tailed datasets. In real-world scenarios, data distributions frequently follow a long-tailed distribution, wherein minority classes are typically inadequately represented. This introduces the notion of "label bias," wherein the classification decision boundary is significantly influenced by the majority classes\cite{yang2020rethinking}.

\begin{figure}[t]
\centerline{\includegraphics[scale=0.3]{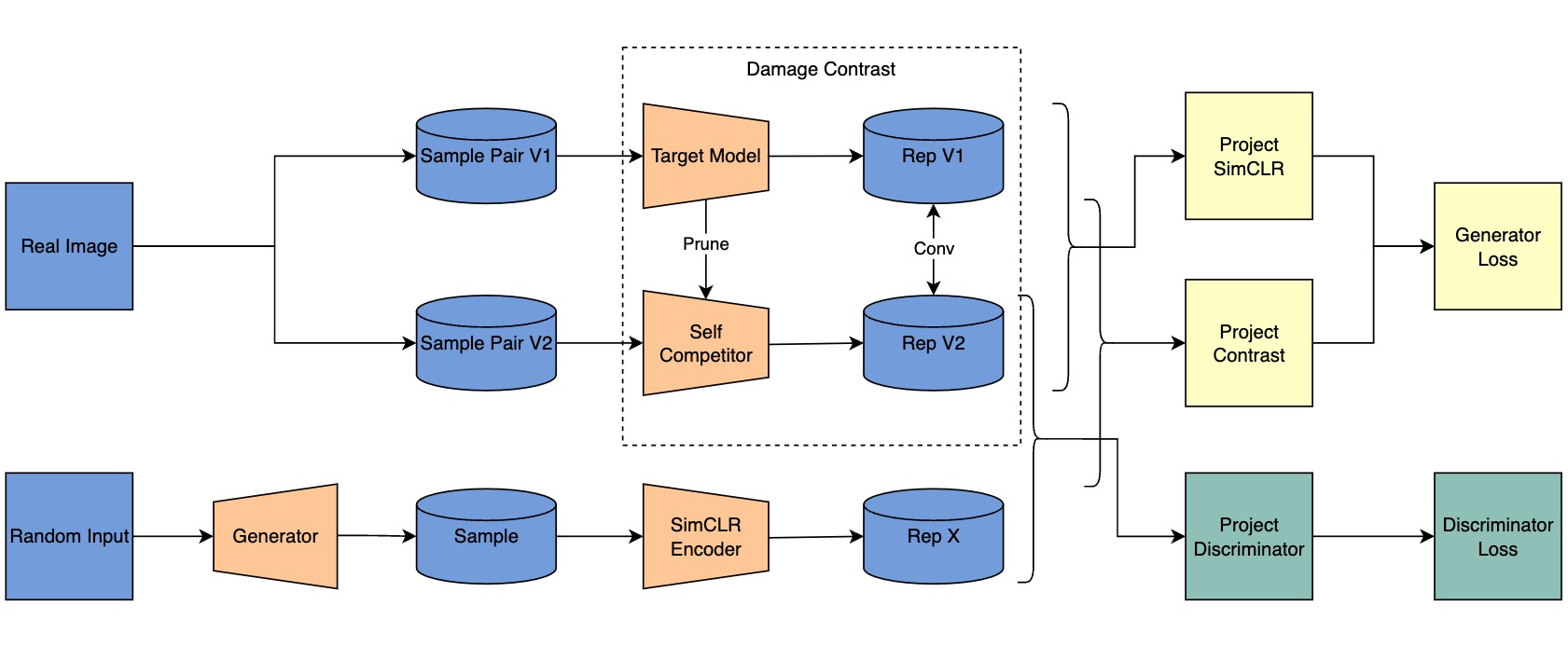}}
\caption{The structure of Damage GAN.}
\label{fig:damage}
\end{figure}

To enhance the performance of ContraD GAN in the context of imbalanced data settings, a potential strategy involves refining the discriminator component. Recent advancements in contrastive learning, drawing inspiration from the SimCLR architecture, have demonstrated superior performance compared to the original model. Consequently, exploring alternative contrastive learning architectures by substituting the SimCLR component within ContraD GAN presents a promising avenue.

An illustrative example of a relevant study addressing imbalanced data within contrastive learning is the self-damaging contrastive learning method (SDCLR)\cite{jiang2021self}. In SDCLR, a branch of the original SimCLR framework is adapted to create a pruned branch, wherein samples from minority classes are treated differently through the assignment of larger losses. This adaptation guides the model to assign heightened importance to samples from the minority classes, effectively mitigating the imbalance issue.

Guided by these insights, we propose the replacement of the SimCLR segment in the original ContraD GAN model with SDCLR (as depicted in Figure \ref{fig:damage}). By integrating a contrastive learning structure that exhibits enhanced performance in imbalanced data scenarios, we anticipate an overall improvement in the GAN framework's performance when handling datasets characterized by imbalances. This modification introduces a promising pathway to overcome challenges linked to imbalanced data within ContraD GAN, extending its utility to real-world applications.

\section{Experimental Results}
\label{sec:experiment}

\subsection{Datasets}
The study utilizes the CIFAR-10 dataset, comprising 60,000 entries with dimensions of 32x32 pixels. To comprehensively evaluate the proposed model, the dataset was employed in three distinct configurations:

\begin{itemize}
\item Full Dataset: This configuration serves as a baseline for comparison, facilitating an assessment of the proposed model's overall performance.

\item Partial Dataset: To delve into the model's efficacy on imbalanced datasets, the original data was reduced to one-fifth of its original size, following the method established by Cui et al.\cite{cui2019class}. This smaller dataset not only possesses an imbalance but also serves as the foundation for controlled experiments.

\item Imbalanced Dataset: Following the generation of the partial dataset, Cui et al. introduced a balance factor to select examples, thereby constructing an imbalanced dataset. For this study, a balance factor of 100 was implemented, resulting in the largest class comprising 4,500 examples, while the smallest class contains only 45 examples. 
\end{itemize}
 
The comparative analysis entails evaluating different GANs on the Full, Partial, and Imbalanced datasets, thus yielding insights into their respective performances across varying dataset configurations.

\subsection{Evaluation Metrics}
This paper employs two primary metrics, namely the Inception Score (IS) and the Fréchet Inception Distance (FID), to gauge the effectiveness of the GANs.

The Inception Score (IS) evaluates the quality of generated images\cite{barratt2018note}. This assessment involves inputting an image into a neural network, specifically Inception-v3\cite{xia2017inception}, and acquiring the output layer probabilities for each category, denoted as $p(y \mid x)$. Here, $x$ signifies the data feature, and $y$ represents the label. The distribution of labels is represented by $p(y)$. The IS is calculated using the following formula:

\begin{equation}\label{eq1}
IS = \exp \left(\mathbb{E}_{x \sim p_{g}} D_{K L}(p(y \mid x) \| p(y))\right)\\
\end{equation}

The primary aim is for the generator to produce diverse images containing meaningful objects, resulting in a low-entropy distribution $p(y)$ and a high-entropy distribution $p(y \mid x)$. A higher Inception Score (IS) indicates superior performance, showcasing a more substantial KL-divergence\cite{bu2018estimation} between these two distributions.

Fréchet Inception Distance (FID) can be perceived as an advancement of the IS metric, as it takes into account not only the quality of samples generated by GANs but also the influence of real data\cite{obukhov2020quality}. FID compares the statistical attributes of generated images with those of real samples by leveraging features from Inception-v3, unlike IS, which directly provides class assignments. The FID is calculated using the subsequent formula:

\begin{equation}\label{eq2}
FID = \|\mu_r - \mu_g\|^2_2 + Trace(\sum_r + \sum_g -2(\sum_r \sum_g)^{1/2})\\
\end{equation}

In this context, $x_r \sim N(\mu_r, \sum_r)$ and $x_g \sim N(\mu_g, \sum_g)$ represent the 2,048-dimensional activations of the Inception-v3 pool3 layer for real and generated samples, respectively. A reduced Fréchet Inception Distance (FID) value indicates improved performance, highlighting a higher similarity between real and generated images. This similarity is quantified by measuring the distance between their activation distributions.

\subsection{Results}

In this section, we present a series of experiments aimed at comparing the performance of three models: DCGAN, ContraD GAN, and our proposed model, Damage GAN. The main objectives of these experiments are as follows:

Reproducing and evaluating results for both DCGAN and ContraD GAN, while simultaneously assessing the performance of our proposed Damage GAN model using the standard CIFAR-10 dataset.

Additionally, generating two subsets from the original CIFAR-10 dataset, each containing 10,000 samples. One subset is balanced, while the other is deliberately imbalanced. By calculating FID and IS metrics across nine different scenarios (combining three GANs and three datasets), we aim to understand the implications of this data modification on performance.

Furthermore, investigating the potential disparities in class distribution when GANs operate on the imbalanced dataset. We also explore whether Damage GAN contributes to improving this distribution.

Lastly, we aim to determine whether Damage GAN has succeeded in enhancing the image quality of the imbalanced dataset. We accomplish this by comparing FID results for the two minor classes (with the largest generated samples) and the two major classes (with the smallest generated samples). The outcomes of the aforementioned experiments are provided below:

\begin{table}
\centering
\addtolength{\leftskip}{-30pt}
\caption{FID and IS results for 3 GANs with 3 datasets.}
\label{tab:9cases}
\centering
\begin{tabular}{cccccccccc}
\hline
\multirow{2}*{Dataset} & \multicolumn{3}{c}{DCGAN} & \multicolumn{3}{c}{ContraD} & \multicolumn{3}{c}{Damage} \\
~ & FID & IS(mean) & IS(std) & FID & IS(mean) & IS(std) & FID & IS(mean) & IS(std) \\
\hline
Full & 25.47 & 7.40 & 0.19 & 10.27 & 9.02 & 0.28 & 11.04 & 8.66 & 0.16 \\
Partial & 40.57 & 6.42 & 0.14 & 16.72 & 7.92 & 0.21 & 18.56 & 8.12 & 0.29 \\ 
Imbalanced & 55.15 & 5.71 & 0.16 & 29.92 & 7.43 & 0.16 & 28.45 & 7.95 & 0.15 \\ 
\hline
\end{tabular}
\end{table}

a) The CIFAR-10 results for DCGAN and ContraD GAN align with previously published findings\cite{jeong2021training}, as observed from Table \ref{tab:9cases}. However, the Damage GAN takes twice as long to run and exhibits slightly inferior performance compared to ContraD GAN.

b) The FID results for all GANs deteriorate when the training sample size is reduced, and the degradation worsens in the case of imbalanced training sets. Among the GANs, the Damage GAN exhibits the lowest rate of deterioration, as depicted in Table \ref{tab:9cases}. Moreover, on the imbalanced dataset, the Damage GAN outperforms the ContraD GAN by 5\% in terms of scores.

\begin{table}
\centering
\addtolength{\leftskip}{-32pt}
\caption{Samples of classes on Partial and Imbalanced datasets.}
\label{tab:sample}
\centering
\begin{tabular}{cccccc cccccc}
\hline
Dataset & air & car & bird & cat & deer & dog & frog & hrs & ship & truck & Total\\
\hline
Partial & 1,116 & 1,116 & 1,116 & 1,116 & 1,116 & 1,116 & 1,116 & 1,116 & 1,116 & 1,116 & 11,160 \\
Imbalanced & 348 & 969 & 125 & 208 & 1,617 & 75 & 4,500 & 581 & 2,697 & 45 & 11,165 \\
\hline
\end{tabular}
\end{table}

\begin{table}
\centering
\addtolength{\leftskip}{-20pt}
\caption{Distribution for classes on Partial and Imbalanced datasets.}
\label{tab:dist}
\centering
\begin{tabular}{cccccc cccccc}
\hline
Dataset & Model & air & car & bird & cat & deer & dog & frog & hrs & ship & truck \\
\hline
\multirow{3}*{Partial}
~ & DCGAN & 0.94 & 0.52 & 1.55 & 0.75 & 1.50 & 0.69 & 1.74 & 0.58 & 1.18 & 0.53 \\
~ & ContraD & 0.96 & 1.23 & 1.10 & 0.90 & 0.85 & 0.74 & 1.11 & 1.13 & 0.93 & 1.02 \\
~ & Damage & 1.02 & 1.22 & 1.14 & 0.75 & 0.87 & 0.79 & 1.08 & 1.14 & 0.93 & 1.02 \\
\hline
\multirow{3}*{Imbalanced}
~ & DCGAN & 0.94 & 0.19 & 4.67 & 1.04 & 0.96 & 0.51 & 0.98 & 0.49 & 0.86 & 1.73 \\
~ & ContraD & 1.35 & 1.05 & 3.32 & 0.85 & 0.75 & 0.89 & 0.81 & 1.11 & 0.83 & 2.11 \\
~ & Damage & 1.16 & 0.94 & 2.85 & 1.05 & 0.75 & 0.64 & 0.85 & 1.17 & 0.85 & 1.42 \\
\hline
\end{tabular}
\end{table}

c) Table \ref{tab:sample},\ref{tab:dist} presents the deviation of the generated samples from the training samples for both Partial and Imbalanced datasets after classifying the generated samples into classes using the linear evaluator. As anticipated, the deviation for the balanced set is close to 1, with a mean value of 1. On the other hand, the imbalanced GANs display a mean deviation greater than 1, indicating under-representation of major classes and over-representation of minor classes. The Damage GAN generates a distribution that closely resembles the generated distribution.

\begin{table}
\caption{FID scores for minor and major classes in the Imbalance dataset.}
\label{tab:fid}
\centering
\begin{tabular}{cccc}
\hline
Model & Total Classes & Major Classes & Minor Classes \\
\hline
DCGAN & 55.15 \\
ContraD & 29.92 & 31.87 & 32.65 \\
Damage & 28.45 & 31.15 & 31.15 \\
\hline
\end{tabular}
\end{table}

d) The FID results for the imbalanced dataset show a 5\% improvement compared to the ContraD GAN score (see Table \ref{tab:fid}). The quality of FID data for the majority and minority classes is impacted by the smaller generated sample size of the minority class, but it exhibits a similar 5\% improvement. In contrast, the majority class experiences a less than 2\% improvement. These results align with the published SDCLR results, which analyze the accuracy of the linear evaluator, the imbalanced dataset, and the minority and majority classes.
The two primary categories in this context are referred to as "frogs" and "ships," while the less common categories are "dogs" and "trucks." Due to the limited number of samples available for the less common categories, a subset of 100 samples is used when calculating FID. It has been observed that FID decreases as the sample size increases, until reaching around 5,000 samples. To determine the FID for both the majority and minority classes, 100 samples were randomly selected from each imbalanced dataset and compared with the full imbalanced datasets (ContraD GAN, Damage GAN) to establish a scale factor. This scale factor, approximately 1/16, was then applied to adjust the FID for the minority and majority classes.
Upon a visual examination of the majority and minority classes, it appears that the quality of the minority classes is inferior to that of the majority classes. However, this discrepancy is not reflected in the FID scores, which raises the need for further investigation. It is possible that the issue is related to the small sample size and warrants additional exploration.

\section{Conclusion}
\label{sec:conclusion}
The primary objective of this paper is to enhance the performance of Generative Adversarial Networks (GANs) when confronted with imbalanced datasets that closely resemble real-world data distributions. In contrast to prior researchers who focused on altering the generator, our approach involves modifying the original GAN by replacing the discriminator with Self-Damaging Contrastive Learning (SDCLR). Comparative analyses between the baselines (ContraD GAN and DCGAN) and our Damage GAN model reveal noteworthy enhancements, particularly in terms of Fréchet Inception Distance (FID) and Inception Score (IS), with a distinct emphasis on the standard deviation of IS. This signifies that our model generates more consistent outputs compared to the baselines. Moreover, the images produced by Damage GAN showcase improved quality for the major classes.

Nonetheless, our study does present certain limitations. Primarily, our experimentation utilized the CIFAR-10 dataset, which comprises relatively small-sized images. Future endeavors should consider evaluating our model on the GTSRB dataset, characterized by larger images. Furthermore, the dataset in our study features mutually exclusive labels, whereas real-world scenarios often involve images with multiple labels. It would be valuable to explore the applicability of our model in generating complex industry images containing diverse elements. Lastly, while Damage GAN effectively balances minor classes and enhances performance for major classes within imbalanced datasets, the FID of the minor classes increases, indicating potentially lower image quality compared to the baselines.

In conclusion, our modification presents improvements for GANs, particularly in the context of addressing imbalanced datasets, and holds promise for future advancements. Further investigations are recommended to gain deeper insights into the factors influencing FID changes across different classes. Additionally, testing the model on datasets featuring larger image sizes and quantities is suggested to validate results obtained from CIFAR-10 and to explore the model's potential in handling high-resolution images.

\bibliographystyle{splncs04}
\bibliography{ae_ref}
\end{document}